\documentclass{new_tlp}
\usepackage{amsthm} 
\usepackage{mathptmx}
\usepackage{svg}
\usepackage{amssymb}
\usepackage{enumitem}
\usepackage{url}
\usepackage{pifont}
\usepackage{graphicx}
\usepackage{amsmath}
\usepackage[T1]{fontenc}
\usepackage[latin1]{inputenc}
\usepackage{tcolorbox}
\tcbuselibrary{skins}
\usepackage{lipsum} %
\usepackage{listings}
\usepackage{soul}
\newtcolorbox{promptbox}[1][]{%
  enhanced jigsaw,
  colback=white, %
  colframe=black, %
  sharp corners, %
  boxrule=0.5pt, %
  #1
}
\usepackage{xspace}
\DeclareMathOperator{\codeif}{\mathtt{:-} }
\newcommand{\asp}[1]{\mbox{$\mathtt{#1}$}}
\newcommand{\myParagraph}[1]{\medskip\noindent\textit{#1}\xspace}

\DeclareMathOperator{\naf}{\;\mathtt{not}\;}

\newcommand{\vs}{\vspace{-0.2cm}}

\newcommand{\answerset}[1]{
  \lbrace$\ForEach{;}{\ifnum\thislevelcount=1 \else ,$ $ \fi \asp{\thislevelitem}}{#1}$\rbrace
}
\newcommand{\oursystem}{{LLM2LAS}\xspace}
\theoremstyle{definition}
\newtheorem{definition}{Definition}
\newtheorem{example}{Example}

\newcommand\babi{bAbI\xspace}

\newcommand\bcmdtab{\noindent\bgroup\tabcolsep=0pt%
  \begin{tabular}{@{}p{10pc}@{}p{20pc}@{}}}
\newcommand\ecmdtab{\end{tabular}\egroup}

\newcommand{\antonio}[1]{{\color{black} #1}}

  \title[QA with LLMs and LAS]
        {Question Answering with LLMs and Learning from Answer Sets}

  \author[] {
         Manuel Borroto \\
         Department of Mathematics and Computer Science, University of Calabria, Italy \\
         \email{manuel.borroto@unical.it}
         \and Katie Gallagher \\
         The University of Chicago, USA \\
         \email{krgallagher@uchicago.edu}
         \and Antonio Ielo \\
         Department of Mathematics and Computer Science, University of Calabria, Italy \\
         \email{antonio.ielo@unical.it}
         \and Irfan Kareem \\
         Department of Mathematics and Computer Science, University of Calabria, Italy \\
         \email{irfan.kareem@unical.it}
         \and Francesco Ricca \\
         Department of Mathematics and Computer Science, University of Calabria, Italy \\
         \email{francesco.ricca@unical.it}
         \and Alessandra Russo \\
         Department of Computing, Imperial College London, UK \\
         \email{a.russo@imperial.ac.uk}
         }

\jdate{March 2003}
\pubyear{2003}
\pagerange{\pageref{firstpage}--\pageref{lastpage}}
\doi{S1471068401001193}

\begin{document}

\label{firstpage}

\maketitle

  \begin{abstract}
Large Language Models (LLMs) excel at understanding natural language but struggle with explicit commonsense reasoning.
A recent trend of research suggests that the combination of LLM with robust symbolic reasoning systems can overcome this problem on story-based question answering tasks. 
In this setting, existing approaches typically depend on human expertise to manually craft the symbolic component. We argue, however, that this component can also be automatically learned from examples.
In this work, we introduce \oursystem, a hybrid system that effectively combines the natural language understanding capabilities of LLMs, the rule induction power of the Learning from Answer Sets (LAS) system ILASP, and the formal reasoning strengths of Answer Set Programming (ASP). LLMs are used to extract semantic structures from text, which ILASP then transforms into interpretable logic rules.
These rules allow an ASP solver to perform precise and consistent reasoning, enabling correct answers to previously unseen questions. Empirical results outline the strengths and weaknesses of our automatic approach for learning and reasoning in a story-based question answering benchmark.
\end{abstract}

  \begin{keywords}
    Logic-based Learning \and Knowledge Representation,\and Question and Answering (Q\&A).
  \end{keywords}

\section{Introduction}
One of the longstanding challenges in Artificial Intelligence (AI) is equipping machines with the ability to perform commonsense reasoning and to learn such knowledge autonomously from experience or text~\cite{DBLP:journals/cacm/DavisM15}. 
This involves not only understanding implicit knowledge about the world but also applying it flexibly to novel situations, an ability that remains difficult for current AI systems. 
Nonetheless, in machine comprehension and question answering (Q\&A) tasks, AI models frequently achieve high performance by exploiting statistical regularities and shallow text patterns rather than through the acquisition of explicit commonsense knowledge or the execution of systematic reasoning processes~\cite{Al-NegheimishMR21}.

Despite their impressive recent successes, LLMs are no exception to the broader limitations of current AI systems. 
They have been shown to exhibit limited reasoning capabilities and often generate unfaithful or incorrect answers~\cite{zheng2023does}, leading to underperformance on benchmarks specifically designed to evaluate natural language reasoning~\cite{DBLP:conf/nips/Wei0SBIXCLZ22,DBLP:journals/corr/abs-2008-01766}.
While recent techniques, such as Chain-of-Thought prompting~\cite{DBLP:conf/nips/Wei0SBIXCLZ22}, problem decomposition, and in-context learning~\cite{zhao2023survey}, suggest that these models can exhibit some reasoning-like behavior, their capabilities remain limited, often relying on implicit pattern matching rather than robust, generalizable reasoning mechanisms~\cite{DBLP:conf/nips/Wei0SBIXCLZ22,DBLP:journals/corr/abs-2008-01766}.
Moreover, the lack of transparency and explainability in LLMs makes it challenging to determine whether they truly acquire and apply commonsense reasoning~\cite{DBLP:conf/acl/SapSBCR20}. 

On the other hand, LLMs have demonstrated strong capabilities in processing and generating natural language text. 
Notably, they have proven effective in semantic parsing, the task of translating natural language sentences into formal representations~\cite{DBLP:conf/iclr/DrozdovSASSCBZ23}. 
This ability positions LLMs as valuable components for bridging the gap between unstructured language and structured, machine-interpretable logic.
Recent neuro-symbolic approaches integrate LLMs into formal reasoning frameworks~\cite{DBLP:journals/aim/Kautz22}, exploiting their effectiveness in translating natural language into structured representation.
This line of research demonstrates that such combinations can address some of the inherent limitations of LLMs, particularly their lack of explicit reasoning and factual reliability, while retaining their strengths in language generation and semantic interpretation.
For instance, it has been shown that the coherence and consistency of LLMs in story completion tasks can be significantly enhanced by combining LLM-based semantic parsing (to translate text into formal representations) with symbolic reasoning systems that evaluate the correctness of the LLM-generated sentences~\cite{DBLP:conf/nips/NyeTTL21}. 
Moreover, Ishay et al. combine LLMs with Answer Set Programming (ASP)~\cite{DBLP:conf/aaai/Lifschitz08,DBLP:journals/cacm/BrewkaET11} to solve logic puzzles~\cite{DBLP:conf/kr/IshayY023}. 
Yang et al. combine GPT3-based semantic parsing with an ASP knowledge module to perform reasoning, showing state-of-the-art performance on several benchmarks~\cite{DBLP:conf/acl/YangI023}. 
{These approaches also demonstrated that ASP, due to its expressive and robust declarative semantics, is a particularly well-suited symbolic formalism for supporting reasoning in neuro-symbolic systems.}
{However, despite} yielding more robust and interpretable reasoning over textual inputs, they typically depend on manually crafted symbolic knowledge for the reasoning component. 
This manual intervention is time consuming, requires substantial domain expertise, and results in an explicit limitation to scalability and generalization across diverse tasks or domains.

We claim that the symbolic component need not be manually specified, but can instead be automatically learned from examples. Through our proposed approach, the feasibility of this direction is demonstrated, showing that meaningful and generalizable symbolic knowledge can be induced from limited supervision.
{More in detail, we develop the ideas of combining ASP with LLMs~\cite{DBLP:conf/kr/IshayY023} for robust reasoning, and introduce \oursystem, a hybrid system that effectively combines the natural language understanding capabilities of LLMs, the rule induction power of the Learning from Answer Sets (LAS) system ILASP~\cite{DBLP:journals/corr/abs-2005-00904}, and the formal reasoning strengths of Answer Set Programming (ASP)~\cite{DBLP:journals/cacm/BrewkaET11}. }

\oursystem integrates an LLM-based semantic parser with ILASP, a system for inductive learning of knowledge in ASP specifications. The semantic parser extracts symbolic representations from natural language stories and questions, which are then used to automatically construct ILASP learning tasks. Given a story, along with associated questions and answers, \oursystem iteratively learns from narratives the underlying commonsense logic rules required to solve the task. The induced knowledge is general and transferable, enabling an ASP system to correctly answer questions about previously unseen texts. 

The key components of \oursystem include:
\begin{itemize}
    \item An open-source LLM-based few-shot semantic parser for generating from natural language both
    \begin{itemize}
        \item [$(i)$] ASP representations of the input stories, and 
        \item [$(ii)$] mode bias declarations to drive learning from answer sets systems%
        \footnote{Mode bias~\cite{DBLP:journals/corr/abs-2005-00904} is a form of syntactic constraint that defines the set of logic rules that the system is allowed to consider when learning, see Section~\ref{sec:prelim:ilasp}.}
    \end{itemize}
    \item A learning module built upon ILASP, designed to induce commonsense knowledge required for answering questions about narrative texts.
    \item A reasoning module for answering questions about a story using the learned common-sense knowledge, which is based on the \texttt{clingo} ASP solver~\cite{DBLP:journals/tplp/GebserKKS19}.  
\end{itemize}

We evaluated our approach on the \babi Question Answering dataset~\cite{DBLP:journals/corr/WestonBCM15}, a widely used benchmark comprising several tasks designed to test various forms of reasoning, including deduction, induction, coreference resolution, and temporal reasoning. 
The empirical evaluation highlights both the strengths and current limitations of our automated approach to learning and reasoning in story-based question answering tasks. 
\oursystem represents a promising step towards the development of more autonomous, interpretable, and robust systems capable of reasoning over natural language inputs.

\section{Related Work}

Mitra et al. developed a three-layer Q\&A system that combines statistical methods with inductive rule learning and reasoning \cite{DBLP:conf/aaai/MitraB16}. The system includes a Statistical Inference layer, which uses an Abstract Meaning Representation (AMR) parser,  a Translation layer, which converts the AMR parser output into Event Calculus syntax using a naive deterministic algorithm, and the 
Reasoner layer, which uses a modified version of the ILP (inductive logic programming~\cite{DBLP:journals/jair/CropperD22}) algorithm XHAIL~\cite{ray2009nonmonotonic} to learn the knowledge required for reasoning.
The system achieves on the \babi dataset an accuracy of 99.68\%, but requires users to manually specify mode declarations and task-dependent background knowledge.

Nye et al. proposed a neuro-symbolic approach to improve the coherence and consistency of text generation in a story completion task~\cite{DBLP:conf/nips/NyeTTL21}. The approach uses GPT-3 to generate candidate completion sentences and an LLM-based parser to derive logical representations of a given story and generated sentences. The latter are compared to symbolic candidates inferred using a minimal world model to check consistency. Only consistent candidates are considered for the final generation.
The system performs well on different benchmarks~\cite{DBLP:journals/corr/WestonBCM15,DBLP:conf/emnlp/SinhaSDPH19,DBLP:conf/nips/RuisABBL20}. 
However, the main limitation is the manual design of the world model, which is task specific.

Ishay et al. combined LLM and ASP to solve logical puzzles in a step-by-step manner~\cite{DBLP:conf/kr/IshayY023}. The method uses GPT-3 with prompt engineering to extract relevant objects, their categories and typed predicates from text descriptions of the puzzles. It then generates an ASP program that captures the rules of the given puzzle, using a Generate-Define-Test approach. The outcomes are computed symbolically using the generated ASP program. The method is interpretable but requires human intervention to resolve errors in the generation process. 

Yang et al. demonstrated GPT-3 to be effective in few-shot semantic parsing of natural language into ASP representation~\cite{DBLP:conf/acl/YangI023}. Their approach handles Q\&A tasks but with task-specific manually handcrafted background knowledge, achieving promising results on different NLP benchmarks, {included the bAbI dataset.}
{The authors also conducted additional experiments to evaluate the capacity of LLMs themselves to handle reasoning tasks.
They used a generation-only approach based on GPT-3.5 and various prompting techniques (i.e., Few-shot and CoT).
The results demonstrated that, while LLMs can achieve decent results on some tasks, their overall performance is significantly lower compared to the proposed neuro-symbolic approach.
}
Our approach differs from this work in that we \emph{learn the relevant knowledge needed to solve a task}.

{Alviano et al., in their first~\cite{DBLP:conf/cilc/AlvianoG24} and second report~\cite{DBLP:conf/kodis/AlvianoSGR24}, introduced the LLM2ASP framework, which integrates the reasoning capabilities of ASP with the natural language processing capabilities of LLMs. They proposed a YAML-based format for specifying prompts and encoding domain-specific background knowledge. In this framework, LLMs process the input prompt to generate relational facts or ground truth, which are reasoned upon using an ASP program. 
The resulting output from the ASP program is converted back into natural language using LLMs to provide a better user experience.
Kim et al.~\citeyear{DBLP:conf/emnlp/0001KY24} addressed the reasoning capabilities of black-box LLMs. They proposed a novel approach called COBB (Correct for improving QA reasoning of Black-Box LLMs). 
The approach utilizes a trained adaptation model to map the often-imperfect reasoning of an initial black-box LLM to the correct reasoning.
The adaptation model is based on an open-source LLM model and trained over a set of representative pairs of correct and incorrect reasoning. 
The proposed approach's effectiveness depends on the quality of training pairs and the capability of open-source LLM. 
In addition, it requires ground-truth human labels to judge the correctness of reasoning, which is a time-consuming task.}

Wu et al.~\citeyear{DBLP:conf/aaai/WuHSCL24} proposed MindMap, {a fully LLM-based} approach to enhance the multi-step reasoning in LLMs by constructing evidence chains of facts associated with a common subject. 
The approach puts the related facts together to prevent missing crucial information.
The chains created by MindMap can be combined with Chain-of-Thought (CoT) and Selection-Inference (SI) to improve the performance in logical reasoning tasks. The framework consists of three main modules, i.e., (i) evidence chain construction, (ii) chain summarization, and (iii) chain utilization for reasoning. The approach was evaluated on a subset of the \babi dataset (tasks 1-3) and the ProofWriter~\cite{DBLP:conf/acl/TafjordDC21} dataset, demonstrating that integrating MindMap with CoT and SI leads to significant improvements.
{Despite these clear improvements, the overall performance remains below that of neuro-symbolic approaches, with hallucinations during inference representing a significant contributing factor.
These results highlights that, despite recent progress, obtaining accurate and consistent reasoning from LLMs remains a challenge.
}

{In addition to the approaches discussed above, there are other neuro-symbolic methods that address similar problems while relying on symbolic formalisms other than ASP, such as Prolog~\cite{DBLP:conf/hopl/ColmerauerR93} and constraint programming~\cite{DBLP:books/daglib/0018273}. Recent surveys~\cite{DBLP:journals/corr/abs-2502-15652,DBLP:journals/corr/abs-2310-00836} provide an up-do-date overview of these neuro-symbolic approaches.}

We adopt Learning from Answer Set (LAS) to learn the knowledge needed to solve a Q\&A task, thus reducing human intervention, and exploit LLM-based semantic parsing capability to automatically generate LAS learning tasks from the given natural language dataset. This combination of LLM and LAS is novel and offers promising performance.

{This paper is an extended and revised version of the conference paper by Kareem et al.~\citeyear{DBLP:conf/lpnmr/KareemGBRR24}. In particular, this paper streamlines the ideas of \citeANP{DBLP:conf/lpnmr/KareemGBRR24} by adopting a simpler workflow, that replaces classic NLP techniques with LLM-based techniques, replacing parts-of-speech algorithms with few-shot prompting to define the learning bias for the tasks. Furthermore, several extensions were introduced in the implementation, ranging from more up-to-date LLMs (LLama 3.3 70B in place of the smaller model Falcon 7B) to smarter caching strategies for LLM outputs and learning tasks' hypothesis space. This extension makes the approach more flexible and expands its applicability. As a result, we are able to solve more tasks from the \babi dataset, that were unfeasible in~\citeNP{DBLP:conf/lpnmr/KareemGBRR24} due to the complexity of fact extraction.
}

\section{Preliminaries}\label{sec:prelim}

This section consists of a brief recap on Answer Set Programming (Section~\ref{sec:prelim:asp}), the Event Calculus formalism to reason about actions (Section~\ref{sec:prelim:eventcalculus}), the inductive logic programming under the Learning from Answer Sets framework (Section~\ref{sec:prelim:ilasp}), and Large Language Models (Section~\ref{sec:prelim:llms}), providing relevant notions that will be referred to throughout the paper.

\subsection{Answer Set Programming}\label{sec:prelim:asp}
Answer Set Programming (ASP) is a well-known paradigm for specifying real-world problems, common-sense knowledge and solving combinatorial optimisation problems~\cite{DBLP:journals/cacm/BrewkaET11,Gelfond1988}. 
We provide here a brief recap of the ASP syntax relevant to this paper, referring the reader to~\cite{Gelfond1988,DBLP:journals/cacm/BrewkaET11,DBLP:journals/tplp/CalimeriFGIKKLM20} for a formal account on ASP syntax and semantics.

\paragraph{Syntax.} Given atoms $\asp{h}$, $\asp{b_1,\ldots,b_n}$,
$\asp{c_1,\ldots,c_m}$,
a \emph{normal rule} is of the form $\asp{h \codeif b_1,\ldots, b_n,}$ \break $\asp{\!\naf c_{1},\ldots,\naf c_{m}}$, 
where $\asp{h}$ is the \emph{head},
$\asp{b_1,\ldots, b_n,}$ $\asp{\naf c_{1},\ldots,\naf c_{m}}$ (collectively) is
the \emph{body} of the rule, and ``$\mathtt{not}$'' represents negation as
failure.  Rules $\asp{\codeif b_1,\ldots, b_n,}$ $\asp{\naf
c_{1},\ldots,\naf c_{m}}$ are called \emph{hard constraints}. 
ASP programs include also \emph{choice rules}. A choice rule is a special type of rule of the form $\asp{l\{h_{1},
\ldots, h_{k}\}u\codeif b_1,\ldots, b_n,}$ $\asp{\naf c_{1},\ldots, \naf
c_{m}}$, where $\asp{l}$ and $\asp{u}$ are integers. 
A variable in a rule is said to be \emph{safe} if it occurs in at least one positive literal (i.e.\ the $\asp{b_i}$'s in the above rule) in the body of the rule. 
In this paper, we assume an ASP program to be a set of normal rules, hard constraints and choice rules. 
The semantics of ASP programs is in terms of stable models (or answer sets)~\cite{Gelfond1988}. 

ASP solvers are capable of constructing solutions to real-world problems from a given ASP program specification of the problem and, where needed, ranking solutions according to optimisation criteria. %

\paragraph{Semantics.} The Herbrand Base of a program $P$, denoted $HB_P$, is the set of variable free (ground)
atoms that can be formed from predicates and constants in $P$.  The subsets of
$HB_P$ are called the (Herbrand) interpretations of $P$.  A ground aggregate $\asp{l\{h_{1}, \ldots,
h_{k}\}u}$ is satisfied by an interpretation $I$ iff $\asp{l}\leq|
I\cap\{\asp{h_{1}, \ldots,h_{k}}\}|\leq \asp{u}$.

As we restrict our ASP programs to sets of normal rules, constraints and
choice rules, we can use the simplified definitions of the \emph{reduct} for
choice rules presented in \cite{simpleReduct}. Given a program $P$ and an
Herbrand interpretation $I \subseteq HB_{P}$, the reduct $P^{I}$ is constructed
from the grounding of $P$ in 4 steps. Firstly, removing rules whose bodies
contain the negation of an atom in $I$; secondly, removing all negative literals
from the remaining rules; thirdly, replacing the head of any constraint, or
any choice rule whose head is not satisfied by $I$ with $\asp{\bot}$ (where
$\asp{\bot}\notin HB_P$); finally, replacing any remaining choice rule
$\asp{l \lbrace h_1,\ldots,h_m\rbrace u\codeif b_1,\ldots,b_n}$ with the set of
rules $\lbrace \asp{h_i \codeif b_1,\ldots,b_n} \mid \asp{h_i} \in I \cap
\lbrace \asp{h_1,\ldots, h_m}\rbrace\rbrace$. Any $I \subseteq HB_{P}$ is an
\emph{answer set} of $P$ if it is the minimal model of the \emph{reduct}
$P^{I}$.  We denote with $AS(P)$ the set of answer sets of a program
$P$.  A program $P$ is said to be satisfiable (resp.
unsatisfiable) if $AS(P)$ is non-empty (resp. empty).

\subsection{Simplified Discrete Event Calculus}\label{sec:prelim:eventcalculus}
\emph{Event Calculus} (EC)~\cite{kowalski1986logic} is a logic-based formalism to reason about actions and their effects. The EC formalization of a subject domain consists of a set of first-order rules that define properties of interest in the domain (``fluents''), and domain-independent rules (``axioms'')  that describe general principles about how such properties evolve, that is when, how they become true or false in a given point in time~\cite{ecexplained}. There exist multiple flavours of Event Calculus. In this paper, we are interested in the \emph{Simplified Discrete Event Calculus (SDEC) ~\cite{sdec}.}

\begin{table}[b]
\centering
\begin{tabular}{p{4.0cm} p{4.5cm}}
\hline \hline

\textbf{Predicate} & \textbf{Meaning} \\ \hline
$\asp{holdsAt(f, t)}$ & The fluent $f$ is true at $t$.  \\ 
$\asp{happensAt(f, t)}$ & The fluent $f$ is observed at $t$.  \\ 
$\asp{initiatedAt(f, t)}$ & The fluent $f$ initiates at $t$.  \\ 
$\asp{terminatedAt(f, t)}$ &  The fluent $f$ ceases at $t$.  \\ \hline \hline
\end{tabular}
\caption{Predicates to model Event Calculus as a normal logic program.}
\label{tab:aspeventcalculus}
\end{table}

SDEC can be elegantly implemented in ASP by means of a normal logic program, using predicates $\asp{holdsAt/2}$, $\asp{initiatedAt/2}$, and $\asp{terminatedAt/2}$. Intuitively, SDEC consists of rules that enable to track (and infer) the truth value of fluents over a finite, discrete, linear representation of time.

The axioms of SDEC can be rendered in ASP according to the rules in Figure~\ref{fig:sdecaspaxioms}, and Table~\ref{tab:aspeventcalculus} reports the informal meaning of such predicates. The $\asp{initiatedAt/2}$ and $\asp{terminatedAt/2}$ predicates are used to define the point in times where an event initiates and terminates. Indeed, different fluents have different initiating and termination conditions. The predicate $\asp{holdsAt/2}$ tracks true fluents at any given time point, with $\asp{holdsAt(f,t)}$ modeling that fluent $\asp{f}$ is true at time $t$.

\begin{figure}
{
\begin{align}
&\asp{holdsAt(F,T+1) \codeif initiatedAt(F,T),time(T).}\\
&\asp{holdsAt(F,T+1) \codeif holdsAt(F,T), not\ terminatedAt(F,T),time(T).}
\end{align}
}
\caption{Simple Discrete Event Calculus axioms as ASP rules.}
\label{fig:sdecaspaxioms}
\end{figure}

\subsubsection{Modeling narratives with Event Calculus}

We provide an example of such ASP-based formalization of \emph{narratives} by means of SDEC. A narrative is an ordered sequence of (natural language) statements that describes an event.

\begin{figure}[t]
\centering
\includegraphics[width=\textwidth]{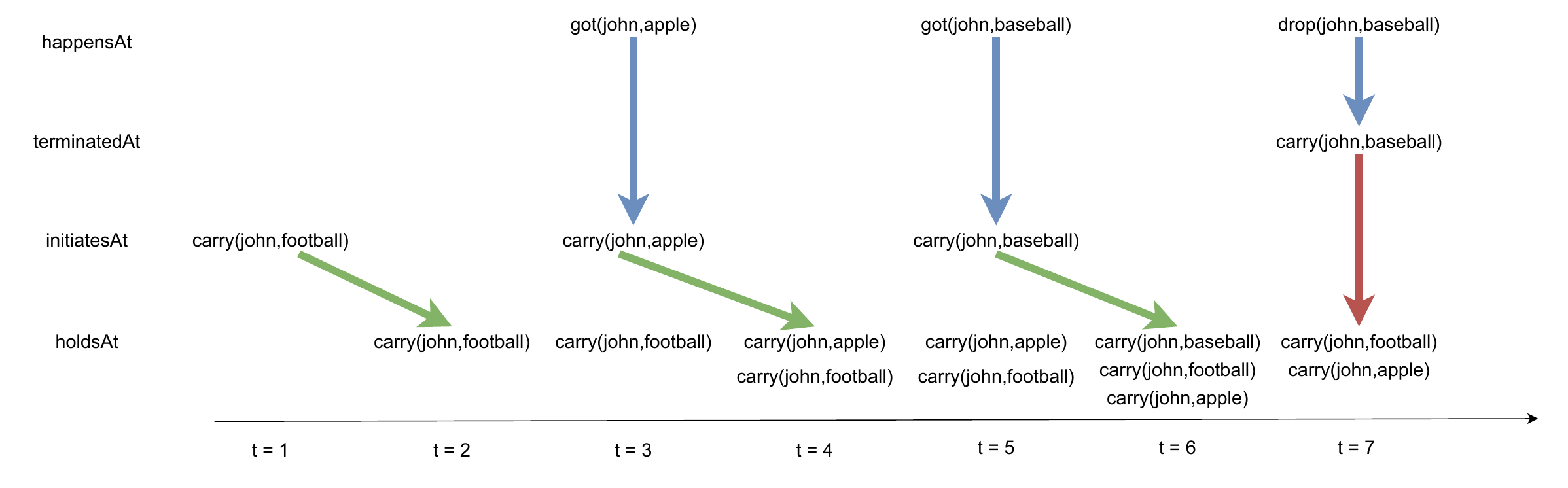}
\caption{Fluents $\asp{carry/2}$ evolving over time, according to SDEC axioms. Narrative's observations---in terms of $\asp{got/2}$, $\asp{drop/2}$ fluents---\emph{trigger} the $\asp{carry/2}$ start/stop (blue arrows), which triggers $\asp{carry/2}$ definitions (green arrows), that dictate truth value over time due to inertia law (``something is true once it initiates and up to the point it terminates''). We can see that John carries with himself the football up to $t=6$ when he drops it; the fluent $\asp{drop(john,football)}$ disables the (default) inertia rule.}
\label{fig:johnsadventures}
\end{figure}

\begin{example}\label{examplesdec}
Consider the following narrative, similar to those in Task 8 of the \babi dataset. Each line consists of a sentence, and we assume that actions that take place in the $i$-th sentence happen at time $i$.
\vspace{0.5cm}
\begin{verbatim}
1. John is carrying the football. 
2. John went to the kitchen. 
3. John got an apple there. 
4. John went to the park. 
5. John got the baseball there. 
6. John dropped the football.

   > What is John carrying?
\end{verbatim}

The narrative involves the agent John, and its action involves interacting with items---picking them up, dropping them--- as well as the moving through several locations.
\end{example}

The narrative provides explicit, point-wise, information about how John interacts with items and moves in space; however, the concept of ``what is John carrying at any given point in time'' is not explicitly provided in the narrative, but is implicit in \emph{what it has been picked up, but not dropped yet}. The first step to model such a narrative in SDEC would be to appropriately choose fluents, and then to provide definitions for its initiating and terminating conditions.

In particular, a possible way to model such scenario is to use the fluent $\asp{got(john,obj)}$ to state that john picks up a given object, and $\asp{drop(john,obj)}$ to state he drops an object. Furthermore, the fluent $\asp{carries(john,obj)}$ states that John is \emph{carrying} a specific item. Indeed, for completeness, one may also wish to include the fluent $\asp{go\_to(john,loc)}$ to state that John is moving to a specific location $\asp{loc}$, however notice that in this particular case, it is not necessary to keep track of John's location to answer the narrative's question. Thus, we can \emph{reify} the narrative by means of the following ASP facts:

$$\asp{initiatedAt(carry(john,football),1).}$$
$$\asp{happensAt(go\_to(john,kitchen),2).}$$
$$\asp{happensAt(got(john,apple),3).}$$
$$\asp{happensAt(go\_to(john,park),4).}$$
$$\asp{happensAt(got(john,baseball),5).}$$
$$\asp{happensAt(drop(john,football),6).}$$

The next step is to provide a \emph{definition} for the fluent $\asp{carry/2}$, that is ``the meaning'' of carrying an object, what determines that John is carrying something with itself and when he stops doing so. Indeed, John starts carrying something with itself once he picks it up, and stops carrying something once he drops it:

$$\asp{initiatedAt(carry(P,O),T) :- happensAt(got(P,O),T).}$$
$$\asp{terminatedAt(carry(P,O),T) :- happensAt(drop(P,O),T).}$$

In this case, the common-sense knowledge that \emph{if someone carries an item he keeps it with itself unless he drops it} is implicit in the inertial law of the second SDEC axiom. {Let $\Pi$ be a logic program that contains the Figure~\ref{fig:sdecaspaxioms} rules, fluents' definitions and the narrative reified onto a set of facts as shown above. Answer sets of $\Pi$ can be partitioned by the second term of each atom (which models time), and we can interpret this model as a \emph{sequence of fluents}, as depicted in Figure~\ref{fig:johnsadventures}. In this case, a single answer set is obtained. However, more complex scenarios (e.g., involving non-deterministic outcomes for actions) can be modeled by means of choice rules and constraints involving the truth value of the fluents, which might yield more than one answer set or no answer sets for the SDEC formalization, which has to be interpreted as \emph{multiple feasible course of actions matching the narrative} or \emph{infeasibility of the narrative (according to SDEC axioms and provided definitions).} Consequently, ASP reasoners can be used to reason about narratives in a more complex way: checking if a given fluent is true at a given point in time, or if a desired fluent is true in all answer sets. These reasoning tasks on narratives would roughly correspond to brave reasoning and cautious reasoning in ASP.}

\subsection{Learning from Answer Sets}\label{sec:prelim:ilasp}
Inductive Logic Programming (ILP)~\cite{DBLP:journals/jair/CropperD22}, which aims at learning logic programs called \emph{hypotheses} that together with an existing background knowledge explain a set of observations, has been extended to learning ASP programs~\cite{law18}. Learning ASP programs allows us to learn a variety of declarative non-monotonic, common-sense theories, including for instance the Event Calculus~\cite{kowalski1986logic} and domain-dependent theories~\cite{katzouris2015incremental}. 
In this paper, we use the \emph{Learning from Answer Sets} (LAS) framework and its state-of-the-art system ILASP ~\cite{law18} for learning ASP programs. The LAS framework solves \emph{learning tasks} which consist of a \emph{background knowledge}, the \emph{mode bias} and a set of \emph{examples}. The background knowledge, denoted as $B$, is an ASP program which describes a set of concepts that are known before learning. 

Formally, the hypothesis space $H$ is defined as a set of (possibly non-ground) rules, and an hypothesis $h$ is a logic program composed of rules in $H$, that is $h \subseteq H$. However, in ILP systems, it is not so common to explicitly provide the hypothesis space, but rather to rely on declarative means to describe it. One possible way to do so in the ILASP system is to provide the hypothesis space by means of \emph{mode biases}.

The mode bias, denoted as $M$ and often called \emph{language bias}, is used to express the ASP programs that can be learned. A \emph{mode bias} is defined as a pair of sets of mode declarations $M=\langle M_{h}, M_{b}\rangle$, where $M_{h}$ (resp.\ $M_{b}$) are called the \emph{head}
(resp.\ {\em body}) {\em mode declarations}. Each mode declaration is a literal
whose abstracted arguments are either $\asp{var(t)}$ or $\asp{const(t)}$, for
some constant $\asp{t}$ (called a \emph{type}). For each type, a set of constants is provided along with the maximum number of variables ($maxv$) that a rule can take, thus constraining the search space induced by $M$. \antonio{In other words, mode biases describe \emph{what} atoms can appear in rules that will describe the hypothesis space; $maxv$ acts as a filter to prune rules that contain more than a given number of variables.} Informally, a literal is
\emph{compatible} with a mode declaration $m$ if it can be constructed by
replacing every instance of $\asp{var(t)}$ in $m$ with a variable of type
$\asp{t}$, and every $\asp{const(t)}$ with a constant of type
$\asp{t}$.

The set of constants of each type is assumed to be given
with a task, together with the maximum number of variables in a rule, giving a
set of variables $\asp{V_1,\ldots,V_{max}}$ that can occur in a hypothesis.
Whenever a variable $\asp{V}$ of type $\asp{t}$ occurs in a rule, the atom $\asp{t(V)}$ is added to the body of the rule to enforce the type. This guarantees the learning of safe rules.
 
\begin{definition} \label{def:mode}
  Given a mode bias $M = \langle M_h, M_b \rangle$, a normal rule $R$ is in the
  hypothesis space $S_M$ if and only if (i) the head of $R$ is compatible with a
  mode declaration in $M_h$; (ii) each body literal of $R$ is compatible with a
  mode declaration in $M_{b}$; and (iii) no variable occurs with two different
  types.
\end{definition}

\begin{example}[ILASP Mode Biases (Normal Rules)]
In the input language of the ILASP system~\cite{DBLP:journals/corr/abs-2005-00904}, mode biases (for normal rules) are provided by means of the \asp{\#modeh} and \asp{\#modeb} directives. Other directives are available to express choice rules or disjunctive rules. As an example, the mode bias:

\begin{verbatim}
#modeh(a). #modeh(b).
#modeb(a). #modeb(b).
\end{verbatim}

\noindent states that the ground atoms \asp{a} and \asp{b} can belong to the head or to the body of a rule. Thus, this can be understood as a compact, declarative specifications for the set of rules\footnote{The output can be obtained by running the command \texttt{ILASP -s bias.lp}, where \texttt{bias.lp} is a file containing the above-specified directives.}:

\begin{center}
\begin{minipage}[t]{0.3\textwidth}
\begin{verbatim}
1 ~  :- a.
1 ~  :- b.
1 ~ b.
1 ~ a.
1 ~  :- not b.
1 ~  :- not a.
2 ~  :- a; b.
2 ~ b :- a.
2 ~ a :- b.
2 ~  :- a; not b.
2 ~ a :- not b.
2 ~  :- b; not a.
2 ~ b :- not a.
2 ~  :- not a; not b.
\end{verbatim}    
\end{minipage}
\end{center}

\noindent where the integer left of the tilde corresponds to the \emph{cost} of the rule, that is the number of literals it contains. Thus, the provided mode biases implicitly define as hypothesis space the set of programs that is obtained by combining the above rules.
\end{example}

The set of examples, denoted as $E$, describes a set of semantic properties that the learned ASP program should satisfy. They are defined in terms of \emph{partial interpretations}. A partial interpretation is a pair of sets of ground atoms 
$\langle e^{inc}, e^{exc}\rangle$, called respectively inclusion and exclusion sets. 
An interpretation $I$ \emph{extends} $e$ iff
$e^{inc} \subseteq I$ and $e^{exc}\cap I = \emptyset$. A ILASP example $ex\in E$ is a \emph{context dependent partial interpretation} (CDPI). This is a tuple $ex =
\langle ex_{id}, ex_{pi}, ex_{ctx}\rangle$, where $ex_{id}$ is an identifier for $ex$, $ex_{pi}$ is a partial interpretation and $ex_{ctx}$ is an ASP program
called a \emph{context}. A CDPI $ex$ is \emph{accepted} by a program $P$ if and only if
there is an answer set of $P\cup ex_{ctx}$ that extends $ex_{pi}$. The idea of a context-dependent example is that each context only applies to a particular example. This is suitable for our question-answering tasks where the answer to a question is normally contextualized with respect to the story or text provided to the learner. 
Formally, an ILASP \emph{context-dependent learning task} is defined as follows.

\begin{definition}\label{def:loas_context} 
A \emph{Context-dependent Learning task} ($ILP_{LAS}^{context}$) is a tuple $T=\langle B, S_M,$ $E\rangle$ where $B$ is an
ASP program, called the background knowledge, $S_{M}$ is the set of rules allowed in the hypotheses (the hypothesis space), and $E$ is a set of CDPIs. A hypothesis $H$ is an inductive solution of $T$ (written $H \in ILP_{LAS}^{context}(T)$) if and only if:

\vspace{-2mm}
  \begin{enumerate}[label=\arabic*., leftmargin=1.5em, itemindent=0em, labelsep=0.5em] 
    \item
      $H \subseteq S_M$;
    \item
      $\forall \langle ex_{id}, ex_{pi}, ex_{ctx}\rangle\in E$,  $\exists A \in AS(B\cup ex_{ctx} \cup H)$ such that $A$ extends $ex_{pi}$.
  \end{enumerate}
\end{definition}

A learning task may have multiple inductive solutions. These are scored in terms of their length (i.e., number of literals they include), $score(H,T) = |H|$. An inductive solution $H\in ILP_{LAS}^{context}(T)$ is \emph{optimal} if there is no other inductive solution $H'\in ILP_{LAS}^{context}(T)$ such that $score(H',T)< score(H,T)$.

\subsection{Large Language Models and POS Tagging}\label{sec:prelim:llms}
The introduction of LLM models, such as GPT and BERT, has revolutionized Natural Language Processing (NLP) by enabling machines to process and generate human language with unprecedented accuracy \cite{DBLP:conf/nips/VaswaniSPUJGKP17}. 
These deep neural network models owe their effectiveness to the transformer-based architectures~\cite{DBLP:conf/nips/VaswaniSPUJGKP17}, which utilize self-attention mechanisms to process and contextualize vast amounts of text. 
Most currently available LLMs have billions of parameters and are trained in a self-supervised way to predict missing tokens or the next token in a given sequence. 
LLMs are usually instructed through text prompts to solve a specific task, such as translating or answering questions. 
They have also been used successfully for semantic parsing, i.e., converting text into a structured format for analysis \cite{DBLP:conf/iclr/DrozdovSASSCBZ23,DBLP:conf/nips/NyeTTL21,DBLP:conf/acl/YangI023}.

Part-of-speech (POS) tagging involves assigning labels to tokens within a text based on their grammatical function, i.e., whether the token is a noun, verb, adjective, adverb, or other \cite{DBLP:books/lib/JurafskyM09}. 
Given a sequence \(x_1, x_2, \ldots, x_n\) of words (tokens) and a set of tags, the task is to generate a sequence \(y_1, y_2, \ldots, y_n\) of tags, where \(y_i\) represents the assigned tag for the input \(x_i\). 
POS tagging presents challenges due to word ambiguities because a word can have multiple meanings and functions depending on the context in which it is used.
In our approach, we employ the spaCy library (\url{https://spacy.io/}).

\begin{figure}[t]
    \centering
\includegraphics[width=0.9\linewidth]{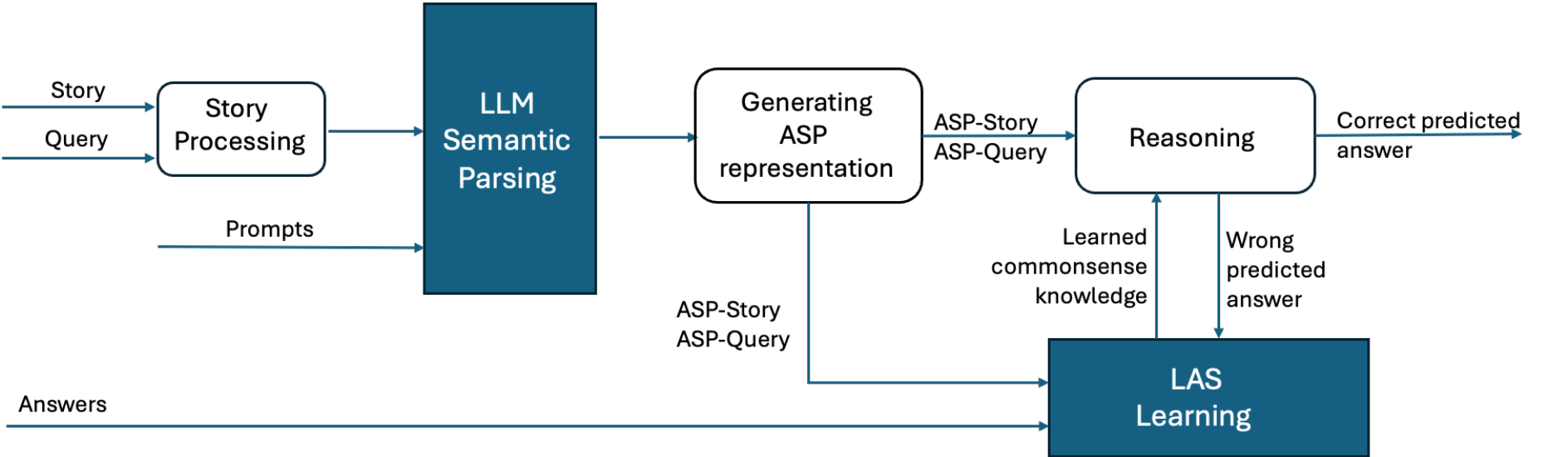}
    
    \caption{Architecture of \oursystem.}
    \label{fig:NetworkOverview}
\end{figure}

\section{Methodology}
In this section, we present our neuro-symbolic system \oursystem, which combines LLMs with LAS to learn commonsense knowledge for story-based Q\&A expressed in natural language.
As illustrated in Figure~\ref{fig:NetworkOverview}, the system consists of several modules.
The \emph{Story Processing} module normalizes the story statements and enriches them with POS tagging data, while the \emph{LLM Semantic Parsing} generates relevant fluent and mode bias representations from the given story. 
These are then used to generate an ASP representation of the narrative described in the story. The reasoner module attempts to answer the question using the extracted narrative and the domain-independent rules given in Figure~\ref{fig:sdecaspaxioms}. If the answer is incorrect, the learner module is invoked to learn relevant commonsense knowledge from the given narrative, question, and ground truth answer. 
In the following, we detail each of these steps.

\myParagraph{Story Processing.} 
The module receives as input a story and a question from which the system is supposed to learn some knowledge. 
A story consists of an ordered set of statements describing a narrative or a scenario, while the question is designed to be answered by exploiting the information in the story. 
Each question is associated with the correct and incorrect answers. 
{All sentences are \emph{normalized} by identifying basic and compound \emph{coreferences}---that is, whether two different expressions refer to the same entity---in the text and replacing them with their corresponding referents. Coreference resolution is automatically performed using the spaCy~\cite{honnibal2020spacy}}.

For example, in basic coreference, the sentence \textit{``Mary went to the store, and she bought food.''} involves replacing the word \textit{``she''} with \textit{``Mary''}.
Moreover, sentences containing negations are identified and flagged to support the following phases.

\myParagraph{LLM Semantic Parsing.} 
LLMs have proven to work well in many NLP tasks, including semantic parsing~\cite{DBLP:conf/iclr/DrozdovSASSCBZ23,DBLP:conf/nips/NyeTTL21,DBLP:conf/acl/YangI023}. 
We exploit this strength and leverage an LLM to parse narratives and questions into fluent-like representation as in Example~\ref{examplesdec}.
The fluent representations support the creation of the EC representations and the mode bias declarations in the next stage.

Most available LLMs are trained on extensive public data, allowing them to achieve reasonable zero-shot generalization on diverse tasks. 
However, these models are not expected to perform as well in domain-specific semantic parsing tasks, where the inductive bias from pre-training is less favorable.
To address this limitation, we used the few-shot prompting technique, which involves giving the model a few task-specific examples within the prompt to help guide its responses~\cite{DBLP:conf/iclr/DrozdovSASSCBZ23}. 
Listing~\ref{lst:prompt} shows an example of the prompt we have designed to ask the LLM to parse the \babi dataset statements.

\begin{lstlisting}[frame=single,caption={Prompt for Fact Extraction}, label={lst:prompt}, basicstyle=\ttfamily\scriptsize,columns=fullflexible,xleftmargin=0.5em,xrightmargin=0.5em,float=tp,]
Please parse the sentence provided below into a first-order logic predicate form. 
The available predicates names are: go_to, be_in.
Sentence: Mary moved to the bathroom.
Semantic parse: go_to(mary,bathroom)
Sentence: John went to the hallway. 
Semantic parse: go_to(john,hallway)
...

Sentence: Where is Daniel? 
Semantic parse: be_in(daniel,V1)

Please, provide just the parsing data using the examples format.
The sentence to parse is:
Sentence: {{sentence}}
Semantic parse:

\end{lstlisting}

For example, if we ask an LLM model to parse the sentence \textit{``Sam moved to the bathroom.''}, using the previous prompt, the result would be: $\asp{go\_to(sam,bathroom)}$.
Our system parses each statement separately, using the same prompt multiple times to give a precise semantic representation in fluent terms. 
Table \ref{tab:ASPrepresentations} (second column) shows a few examples of fluent representations.

\begin{table}
\scriptsize
\centering
\caption{Examples of statements with fluent, and EC representations (Rep.).}
\begin{tabular}{p{3.5cm} p{3.2cm} p{4.2cm}}

\hline \hline
\textbf{Statement} & \textbf{Fluent Rep.} & \textbf{Event Calculus Rep.} \\ \hline 
Mary went to the garden. &$\asp{go\_to(mary,garden)}$ & $\asp{happensAt(go\_to(mary,garden),T)}$ \\ 

John and Helen went to the store. & $\asp{go\_to(john,store),}$ $\asp{go\_to(helen,store)}$ &  $\asp{happensAt(go\_to(john,store),T)}$, $\asp{happensAt(go\_to(helen,store),T)}$ \\ 

John is in the park or garden. & $\asp{\{be\_in(john,park),}$ $\asp{be\_in(john,garden)\}}$ & 
$\asp{\{initiatedAt(be\_in(john,park),T),}$ $\asp{initiatedAt(be\_in(john,garden),T)\}}$ \\ 

Yesterday Ana went to the park. & $\asp{go\_to(ana,park,yesterday)}$ & $\asp{happensAt(go\_to(ana, park,yesterday),T)}$ \\ \hline \hline
\end{tabular}
\label{tab:ASPrepresentations}
\end{table}

\begin{table}[b]
\centering
\caption{Sentences, fluent representations, and mode bias fluents for a short story.}
\begin{tabular}{p{3.5cm} p{4.0cm} p{4.5cm}}
\hline \hline

\textbf{Sentence} & \textbf{Fluent Representation} & \textbf{Mode Bias Fluents} \\ \hline
Mary is a mouse. & $\asp{be(mary,mouse)}$ & $\asp{be(var(nnp),var(nn))}$  \\ 
Mice are afraid of wolves. & $\asp{be\_afraid\_of(mouse,wolf)}$ & $\asp{be\_afraid\_of(var(nn),var(nn))}$  \\ 
What is Mary afraid of? & $\asp{be\_afraid\_of(mary,V1)}$ &  $\asp{be\_afraid\_of(var(nnp),var(nn))}$  \\ \hline \hline
Jason is thirsty. & $\asp{be(jason,thirsty)}$ & $\asp{be(var(nnp),const(jj))}$  \\ 
Jason went to the kitchen. & $\asp{go\_to(jason,kitchen)}$ & $\asp{go\_to(var(nnp),var(nn))}$  \\ 
Why did Jason go to the kitchen? & $\asp{go\_to(jason,kitchen,V1)}$ &  $\asp{go\_to(var(nnp),var(nn)),const(jj))}$  \\ 
\hline\hline
\end{tabular}
\vspace{-2\baselineskip}
\label{tab:ModeBiasFluents}
\end{table}

\myParagraph{Mode Bias Generation.}
{
Mode bias fluents consist of atoms from the sentence's fluent representation where all arguments have been replaced by their types wrapped in either \textit{``var''} or \textit{``const''}.
The argument types are determined using the POS tagging data and the WH-determiners of the questions. 
If the sentence fluent contains an argument that is a variable (i.e., the sentence is a WH-question), then the variable is given a type in the following way: if the sentence is a \textit{``what''}, \textit{``when''}, or \textit{``where''} question, then the variable's type is \textit{``nn''}; if the sentence is a \textit{``who''} question, then the variable's type is \textit{``nnp''}; if the sentence is a \textit{``why''} question, then the variable's type is \textit{``jj''} (which stands for \emph{adjective}), and if the sentence is a \textit{``how many''} question, then the variable's type is \textit{``number''}. In all other cases, the argument's type is given by its associated POS tag.
The types for all arguments that have a temporal aspect and the types of variables in \textit{``why''} questions have \textit{``const''} wrappings.
The types of all other arguments are given \textit{``var''} wrappings. 
The mode bias fluents aid the learner in automatically generating mode bias declarations for both formal representations. Table~\ref{tab:ModeBiasFluents} provides an example for two narratives.}

{
To handle this task, we introduced an LLM-based semantic parser to generate the mode bias fluents given a sentence and its fluent representation. 
In particular, we designed a prompt that captures the mode bias generation methodology discussed earlier and used it to request the parsing from the LLM, specifically Llama-3.3 70B.
The prompts are available in the following Github repository: \url{https://github.com/IrfanKareem/llm2las/tree/journal}.
This choice allowed us to overcome one of the main limitations of the previous version of the approach~\cite{DBLP:conf/lpnmr/KareemGBRR24}, that relied on spaCy and was not general in generating the mode bias for several tasks.}

\myParagraph{Generating ASP Representation.} 
{Once statements of the story are parsed into their corresponding fluent representation, the next step is to create the EC representations (if needed).}
The EC representation depicts the actions and their effects in the story and comprises the four predicates introduced in Section \ref{sec:prelim}.
The construction of the EC representation from the fluent representation involves choosing an EC predicate and a time point. 
Given a sentence and its fluent representation, we select the predicate according to the following schema: (i) if the sentence is a question, then the \asp{holdsAt/2} predicate is used; (ii) if the base of the literal's predicate is ``be'' and the statement is negated, then the \asp{terminatedAt/2} predicate is used; (iii) if the base of the literal's predicate is ``be'' and the statement is not negated, then the \asp{initiatedAt/2} predicate is used; (iv) otherwise, the \asp{happensAt/2} predicate is used.
The time point for the EC predicate is determined by the sentence's placement within the story.
The first sentence is given time point 1, and every subsequent sentence has a time point determined by the previous one plus 1.
Questions are given a time point according to when they are asked.
Table \ref{tab:ASPrepresentations} shows some examples of statements and their representations.

\myParagraph{Reasoning.}
The reasoning module attempts to answer a question using the information extracted from the story and the learned hypothesis.
It involves automatically generating and solving an ASP program that combines the ASP representations and learned hypothesis.
Ideally, the correct solution to this program (i.e., the answer sets) will contain the correct answer to the question.
To extract the answer from the reasoning output, we divide the questions into two types: \textit{``yes/no/maybe''} and others.
In the case of the former, we use a \textit{representation search} that checks the question's representation against the answer sets based on the following criteria: 
(i) if there is at least one answer set, and the representation is in all answer sets, then return \textit{``yes''}; (ii) if there is at least one answer set, and the representation is in some, but not all answer sets, then return \textit{``maybe''}; (iii) otherwise the answer is \textit{``no''}.
For all other questions, we extract the answer using a \textit{unification search}, i.e., by finding all ground atoms in the set of answer sets that unify with the question's formal representation.
To identify these unifications, a regular expression is constructed from the question's formal representation, replacing variables with the wildcard expression \textit{``.*''}.
Once unifications are detected, the ground terms corresponding to the \textit{``.*''} sections of the regular expressions are added to the answer list. 
For example, the question in Table \ref{tab:ModeBiasFluents} generates 
``$be\_afraid\_of(mary,.*)$``.
In case of a wrong predicted answer, the learner is invoked with the question and correct answer to learn from.

\myParagraph{LAS Learning.}
 Learning commonsense knowledge from story-based Q\&A is initiated through the LAS Learning module. It takes as input the EC representations of the story and the question--generated by the ASP representation module--and the correct and incorrect answers for the question. It creates the context dependent learning task for ILASP by automatically generating the mode bias declarations, using the mode bias fluent representations, and the set $E$ of CDPI examples.
To create the mode bias declarations, the system checks whether the sentence is a question, or whether the base of its fluent predicate is ``be''. 
This two-check scheme suffices to generate the language bias, given our basic sentences and limited \babi dataset vocabulary.
For questions, the system aims to learn the concept introduced in it. 
So, its mode bias fluent representation becomes the argument of a mode head declaration, denoted in ILASP as
\textit{``modeh''}.
If the sentence is not a question and the base of its fluent representation is ``be'', then ILASP \textit{``modeb''} declarations are generated with the sentence's fluent as argument of mode body declaration. For the story presented in Table \ref{tab:ModeBiasFluents} the mode bias are:

\vspace{2pt}
\begin{small}\vs
    \begin{verbatim}
    #modeb(be(var(nnp),var(nn))).
    #modeb(be_afraid_of(var(nn),var(nn))).
    #modeh(be_afraid_of(var(nnp),var(nn))).
    \end{verbatim} 
\end{small}\vs\vs
When \oursystem detects that the task requires reasoning about events,
the language bias, referred to as EC mode bias, includes dedicated predicates. To learn the concept introduced by the question, \oursystem learns if it is \textit{initiated} or \textit{terminated}, considering also the initiation or termination of other fluents in the story.
For the question's fluent, two mode head predicates are generated: \textit{``initiatedAt''} and \textit{``terminatedAt''} and declared as arguments of ILASP \textit{``modeh''}.
A body predicate is created by enclosing the question's fluent in a \textit{``holdsAt"} predicate, which is then wrapped in \textit{``modeb"}.
For sentences that are not questions, the system detects if they describe an initiated state or a state that holds. 
In the first case a mode body predicate is created by enclosing the sentence's fluent into a \textit{``initiatedAt''} predicate, in the second case the sentence's fluent is enclosed into a \textit{``holdsAt''} predicate. Both become arguments of ILASP \textit{``modeb''} declarations.
The EC mode bias declarations for the example in Table \ref{tab:ModeBiasFluents} are as follows: 
\begin{small}
    \begin{verbatim}
    #modeb(initiatedAt(be(var(nnp),var(nn)),var(time))).
    #modeb(initiatedAt(be_afraid_of(var(nn),var(nn)),var(time))).
    #modeb(holdsAt(be(var(nnp),var(nn)),var(time))).
    #modeb(holdsAt(be_afraid_of(var(nn),var(nn)),var(time))).
    #modeh(initiatedAt(be_afraid_of(var(nnp),var(nn)),var(time))).
    #modeh(terminatedAt(be_afraid_of(var(nnp),var(nn)),var(time))).
    #modeb(holdsAt(be_afraid_of(var(nnp),var(nn)),var(time))).
    \end{verbatim}
\end{small}\vs

The formal representation of non-question sentences is used to prove or disprove other facts using the domain-independent EC rules in Figure~\ref{fig:sdecaspaxioms} and the learned hypothesis.
Sentences where ``be''-based verbs do not appear denote actions at a specific time point in the story. 
Thus, their mode bias fluent representation becomes an argument of the \textit{``happensAt''} predicate.
Consider the sentence \textit{``Mary goes to the store.''}, whose mode bias fluent is $\asp{go\_to(var(nnp), var(nn))}$. The system generates %
the mode body declaration:
\begin{small}
    \begin{verbatim}    #modeb(happensAt(go_to(var(nnp),var(nn)),var(time))).
    \end{verbatim} \vs
\end{small} \vs

Another key step is the automatic generation of CDPI examples. Examples are created from questions, stories, and their correct and incorrect answers. Each example corresponds to an incorrectly answered question by the reasoning module, as this would trigger the learning module. Intuitively, the representations of all sentences prior to the question would form the example's context, the formal representations of the correct answer would be part the example's inclusion set and the formal representations of some of the incorrect answers would be part of the example's exclusion set. 

{We distinguish two cases, based on whether the formal presentation of the story uses choice rules (that might lead to multiple answer sets).
\paragraph{No choice rules.} In this case, only positive examples are created. The question's answer defines the example's inclusion and exclusion set: if the question answer is ``yes'', the question representation composes the inclusion set, otherwise it composes exclusion set. In all other cases (that is, questions that are not yes/no) the inclusion set is composed with the question's correct answer, and the exclusion set is populated with a wrong answer's fluent representation.
\paragraph{Choice rules.} If choice rules are present in the formal representation of a story, the example generation has to take into account brave and cautious entailment. For a yes/no/maybe question, if the answer is ``maybe'', then the example will include the question representation in its inclusion set to guarantee that the question's concept occurs in at least one answer set. If the correct answer is a ``yes'', then a negative example is created where the inclusion set is empty and the exclusion set includes the question's correct answers. This is to guarantee that the question's formal representation is true in all answer sets. In all other cases, a negative example is created with an empty exclusion set and inclusion set given by the representation of the question's answers. This is to guarantee the wrong answer will be false in all answer sets.}

To illustrate some of the cases explained above, consider the following story: \textit{Daniel went to the kitchen. Daniel went to the bedroom.} 
The question: \textit{Where is Daniel?} The correct answer is the \textit{bedroom} and incorrect answer is \textit{kitchen}. If the incorrect answer is predicted, then the following example is created: 

\begin{small}
    \begin{verbatim}
    holdsAt(F,T+1) :- initiatedAt(F,T),time(T).
    holdsAt(F,T+1) :- holdsAt(F,T), not terminatedAt(F,T),time(T).

    #modeb(happensAt(go_to(var(nnp),var(nn)),var(time))).
    #modeb(holdsAt(be(var(nnp),var(nn)),var(time))).
    #modeh(initiatedAt(be(var(nnp),var(nn)),var(time))).
    #modeh(terminatedAt(be(var(nnp),var(nn)),var(time))).

    #pos({holdsAt(be(daniel,bedroom),3)},{holdsAt(be(daniel,kitchen)
    ,3)},{time(1..3). happensAt(go_to(daniel,kitchen),1).
    happensAt(go_to(daniel,bedroom),2).}).
    \end{verbatim} \vs
\end{small} \vs \vs 

The system requires minimal background knowledge. If the EC is required, the background knowledge will consist of the rules in Figure~\ref{fig:sdecaspaxioms}, that encode the notion of \emph{inertia} (e.g., if a has initiated previously before time $t$, and has not been terminated, it continues to hold at time $t+1$). 
When the ILASP system is run to solve the generated learning task, if the task is satisfiable, the reasoner is updated with the learned hypothesis.

\section{Empirical Evaluation}
In this section, we report on our evaluation of the proposed approach on a well-known question answering dataset.
To this end, we first describe the \babi dataset from Facebook Research \cite{DBLP:journals/corr/WestonBCM15}, and then provide a description of the hardware and software configurations we have employed and of our baselines. Finally, we comment on the results that confirm the efficacy of our system.

\subsection{Experiment setup}

\myParagraph{Dataset.}
The \babi dataset is composed of 20 non trivial tasks of text understanding and reasoning that was proposed by Facebook Research. The \babi dataset was conceived as a benchmark for assessing a range of natural language reasoning abilities, including deduction, path finding, spatial reasoning, and counting~\cite{DBLP:journals/corr/WestonBCM15}.
Each task of the dataset is constructed by simulating words that represent entities and actions. An entity, denoted as a noun, can be a location, an object, or a person, and possesses internal attributes such as size, color, or relative position to cardinal directions. Within this simulation, each entity can perform ten fundamental actions, with each action associated with a collection of replacement synonyms, pronouns, and temporal adverbs, ensuring lexical diversity within the tasks.
More in detail, the dataset comprises 4 actors, 6 locations, and 3 objects per task, it features stories ranging from 3 to 229 sentences and 1 to 12 questions. Each sentence within a story is uniquely identified and accompanied by its answer for each task. The dataset includes both training and test data for each task, with a strong focus on learning from a few examples. The stories are also available in human-readable formats in various languages. 
In the experiment, we place our focus on the natural language, examining the tasks for which our implementation can be applied.
The considered tasks are detailed in Table~\ref{tab:babiprogress}. 

\myParagraph{Hardware and Software Setup.}
All experiments are conducted on a computer equipped with a AMD EPYC 7313 16-Core processor, 2 TB of RAM, and GPU AMD Instinct MI210 with 64 GB of memory. The experiment pipelines are implemented in the Python programming language version 3.9. 
Our architecture has been implemented using 
the open-source LLM LLama-3.3 70b, Clingo 5.6.2 for reasoning on ASP programs, and ILASP 4.4.1 for learning from answer sets. In particular, we use the \asp{2i} version of the ILASP system, which has proved to be the most suitable for our purposes due to the incremental processing of examples. Concerning the learning parameters, the maximum penalty for the size of the hypothesis was set to 50, and the maximum number of variables was set to 3 for the tasks solved with fluent representation, and to 4 for the tasks solved with EC representation. Each task was evaluated on 1000 training examples\footnote{The values for these parameters has been determined experimentally. Tasks that use the EC require one extra variable due to the \emph{time} variable that appears in the background knowledge.}.
For each considered task, we measure the accuracy (e.g., ratio over correct answers) of compared methods.

Our implementation has been also compared against two baselines from the literature, also based on logic programming: the ILP-based system in {Mitra et al.}~\cite{DBLP:conf/aaai/MitraB16}, and the approach proposed by {Yang et al.}~\cite{DBLP:conf/acl/YangI023}.

\medskip
All the material needed to reproduce our experiments can be downloaded from: {   \url{https://github.com/IrfanKareem/llm2las/tree/journal}}.

\begin{table}[b]\centering
\caption{Tasks of the \babi dataset. ``Solved w/t'' stands for ``solved with impr.''.}
\label{tab:babiprogress}
\begin{tabular}{lccp{0.3\linewidth}}
\hline \hline
\textbf{Task Name} & \textbf{Status} & \textbf{Requires EC?} & \textbf{Notes}\\
\hline
(1) Single Supporting Fact & Solved &$\checkmark$&\\
(2) Two Supporting Facts & Unsolved & & Additional mode bias declarations required and large 
hypothesis space.\\
(3) Three Supporting Facts & Unsolved & & Additional mode bias declarations required and large hypothesis space.\\
(4) Two Argument Relations & Solved & &\\
(5) Three Argument Relations & Unsolved & & ILASP hypothesis space too large.\\
(6) Yes/No Questions & Solved &$\checkmark$&\\
(7) Counting & Solved w/t &$\checkmark$& Counting background knowledge added.\\
(8) Lists/Sets & Solved &$\checkmark$&\\
(9) Simple Negation & Solved &$\checkmark$&\\
(10) Indefinite Knowledge & Solved &$\checkmark$&\\
(11) Basic Coreference & Solved &$\checkmark$&\\
(12) Conjunction & Solved &$\checkmark$&\\
(13) Compound Coreference & Solved &$\checkmark$&\\
(14) Time Reasoning & Solved & &\\
(15) Basic Deduction & Solved &$\checkmark$&\\
(16) Basic Induction & Solved & &\\
(17) Positional Reasoning & Solved w/t & & Recursion removed from learned program.\\
(18) Size Reasoning & Solved & &\\
(19) Path Finding & Solved & &\\
(20) Agents Motivations & Solved & &\\
\hline  \hline 
\end{tabular}
\end{table}

\subsection{Results and discussion}

We discuss our results in three separate paragraphs, considering three different settings.
The first paragraph assesses the system applying --without any expert knowledge intervention-- the workflow in Figure~\ref{fig:NetworkOverview}; the second describes some techniques we applied to the basic workflow to improve its performance; the third paragraph focuses on identifying the cases where our system had some difficulty; finally, we compare it with alternative solutions. The section concludes with a general discussion summarizing our findings.

Table~\ref{tab:babiprogress} summarizes the results obtained while approaching the various tasks in the dataset. 
Each row in the table reports the task number and name, the status of the task, a flag indicating whether the Event Calculus background knowledge is needed, and a short note about improvements made to the basic pipeline (if any).

\myParagraph{Tasks solved within the framework.}
First of all, we report that \oursystem could be applied to all \babi tasks, correctly generating ASP representation for stories and mode biases.
{Then, the system was able to learn in a reasonable time (within 24 hours) the ASP specification for 15 tasks out of 20. In particular, the solved tasks required on average 40 seconds, for a cumulative learning time of 9 minutes.}
The \babi tasks that were fully solved are: 1, 4, 6, 8, 9, 10, 11, 12, 13, 14, 15, 16, 18, 19, and 20. Of these 4, 14, 16, 18, 19, and 20 required no background knowledge, and the remaining, namely 1, 6, 8, 9, 10, 11, 12, 13, and 15, required EC background knowledge. 
\oursystem achieved perfect accuracy (100\% accuracy), i.e., it was able to learn an ASP program solving task without any human intervention.
The average runtime for learning is a matter of a few seconds, once ILASP generates the hypothesis space for the first time. Indeed, the hypothesis space is cached and reused across multiple examples in an incremental learning setup, significantly improving efficiency.
 
\paragraph{Mitigating hypothesis space size.}
\oursystem struggled in the remaining 5 tasks because the ILP task was too expensive, often because of the size of the hypothesis space. 
{Thus, upon inspecting its cause, we applied some ad-hoc improvements on a task-by-task basis to prune the size of the hypothesis, such as adding background knowledge involving aggregates for arithmetic-related reasoning, learning non-recursive programs and marking some predicates as being symmetric or anti-symmetric.} In this way we solved \babi Tasks 7 and 17 with an intervention on the learning tasks. 

Task 7 involves basic arithmetic reasoning, specifically the ability to track the number of items an individual is carrying based on a sequence of actions described in the narrative. 
Here, it is required to learn how to count items, but ILASP cannot learn effectively programs that use aggregates. Nonetheless, counting can be considered basic knowledge, so we added the following rule to the background knowledge:
$$\asp{carriedItems(X,N,T) \codeif holdsAt(carry(X,\_),T), N = \#count\{Z: holdsAt(carry(X,Z),T)\}}.$$ 
\noindent that is, we provide an explicit definition for the ``number of items carried by a person at a given point in time''. Including this rule makes the task solvable by our system, which is able to learn the initiating and terminating conditions for the $\asp{carry/2}$ fluent. {Thus, we obtained 100\% accuracy for this task, with a learning time of around 46 seconds. }

On the other hand, Task 17 deals with positional reasoning and, in particular, the relative positioning of objects in a scene, for example learning the definition of ``being left of something'', and ``right of something'', modeled by means of $\asp{be\_right\_of/2}$, $\asp{be\_left\_of/2}$, $\asp{be\_above\_of/2}$, $\asp{be\_below\_of/2}$ atoms. 
Informally, the learning task involves acquiring both the knowledge that spatial relations such as left-right and above-below are opposites and ``symmetric'', as well as learning rule pairs that define the transitive closure of these spatial predicates.
As an example, the for $\asp{be\_above\_of/2}$ predicate we should have:

\begin{verbatim}
be_above_of(X,Y) :- be_below_of(Y,X).
be_above_of(X,Y) :- be_above_of(X,Z), be_above_of(Z,Y).
\end{verbatim}

We observed that the learning task can be made less heavy by reformulating it so that a non-recursive solution is admitted.
This can be obtained by introducing auxiliary predicates (limited to the head of rules) of the form $\asp{be\_\ast/2}$ for each $\asp{be\_\ast\_of/2}$ predicate.  

Although with this improvement we managed to learn an ASP specification that covers most of the examples, thus circumventing the learning bottleneck, this solution does not generalize as the intended (recursive) solution. {In particular, the system obtained an accuracy of 97.8\%, with a learning time of 16 minutes on average.}

\paragraph{Challenges and open tasks.}
For the remaining tasks, namely 2, 3, and 5, there was no space for reducing the impact of the learning phase. 

Task 5 consists of learning narratives that involve tracking movement, location and possession of objects over time. It is characterized both by the requirement of EC background knowledge, and the need for ternary fluents 
($\asp{give\_to(P_1, P_2, O)}$--$P_1$ has given object $O$ to $P_2$, $\asp{receive(P_1, P_2, O)}$--$P_1$ has received object $O$ from $P_2$). 
\oursystem correctly generates the learning task, but  the joint presence of ternary fluents and EC yields a too large hypothesis space, that the ILASP system is unable to ground (i.e., the ILASP system could not build the set of rules that can appear in an hypothesis) in a reasonable time.

Task 2 consists of narratives that involve tracking the position of objects over time; and Task 3 is a more complex version of Task 2. 
In these cases, the primary challenge arises from the need to learn multiple commonsense notions, such as understanding that an agent is ``in'' a location after moving, or that receiving a gift implies ``possession'' or ``ownership'', that are not explicitly stated within the narrative. 
Being able to process this learning task without substantial additions to the background knowledge remains an open problem.

\paragraph{Comparison with other systems.}
We now compare \oursystem with existing approaches in the literature both in terms of accuracy and in terms of the need for human intervention on the 17 successfully-solved tasks.
For the systems by ~\cite{DBLP:conf/aaai/MitraB16,DBLP:conf/acl/YangI023} we use the accuracy reported in the respective publications.
The approach of \citeANP{DBLP:conf/acl/YangI023}, which relies on humanly-devised ASP programs, reaches 100\% accuracy in all 17 tasks.
On the other hand, the approach by \citeANP{DBLP:conf/aaai/MitraB16} can obtain 100\% accuracy in all tasks but task 16 (where accuracy is of 93.6\%).
Finally, \oursystem achieves 100\% accuracy in all tasks but task 17 where it achieves 97.8\% of accuracy.

Although the performance of the compared methods are essentially aligned in terms of accuracy, there is a major difference that has to be outlined: our approach does not require writing ASP code, which is learned automatically from the examples in the training sets of the \babi dataset. 

\paragraph{Discussion.}
The results reported above show that \oursystem successfully learns, reasons and provides answers over 17 commonsense-driven tasks in the \babi dataset, matching the performance of most of the human-expert manually engineered ASP programs.
Although this is a promising result in combining LLMs with logic reasoners, the experiment also helped to identify two open problems: $(i)$ there are tasks we can in principle solve, but the ILP task is out of reach for the learning system (due to size of generated hypothesis space); and $(ii)$ the approach struggles to deal with notions that are not explicitly mentioned within the narrative. 

Focusing on issue $(ii)$, we observe that while there is a growing consensus that LLMs encode a certain degree of common sense knowledge about the world, the framework proposed in this work does not currently leverage this capability to pre-populate the background knowledge with a set of task-relevant rules. Enabling such integration represents a promising direction for future research, aligning well with current trends in neuro-symbolic learning and the broader effort to bridge statistical and symbolic reasoning.

\section{Conclusion}\label{sec:conclusion}
This work presents \oursystem, a novel hybrid framework that advances the integration of LLMs with symbolic reasoning, by introducing an automated pipeline for learning commonsense knowledge from examples. 
Building on prior research that combines LLMs with symbolic components for story-based question answering, our approach moves a step forward existing methods by eliminating the need for manually crafted logic rules. 
This is obtained by leveraging LLMs for semantic parsing, ILASP for rule induction, and ASP for reasoning.

Our results on the \babi dataset demonstrate that it is not only feasible but also effective (in terms of accuracy) to automatically induce ASP specifications with minimal supervision, thus reducing the reliance on human modeling expertise. 
From an accuracy standpoint, \oursystem is capable of matching solutions based on manually crafted ASP encodings.
At the same time, our findings highlight limitations of our implementation: current Learning from Answer Sets (LAS) systems, such as ILASP, sometimes struggle with scalability when faced with large or complex hypothesis spaces, and with notions that are not explicitly mentioned in the datasets. {Another limitation lies in the support of mathematical reasoning constructs, as at the time being ILASP cannot learn programs with aggregates, whereas LLMs are able to extract this kind of information in the semantic parsing step.}
However, knowledge engineers can mitigate these issues by providing more background knowledge, or by resorting to hypothesis space pruning techniques.

Future works will explore methods to further address this bottleneck, potentially through hypothesis space pruning techniques, automated background knowledge extraction, {support for richer mathematical reasoning constructs (e.g., aggregates)} and tighter LLM-LAS integration, as well as alternative learning strategies.
Overall, \oursystem contributes a promising step towards more autonomous, interpretable, and robust systems for reasoning in natural language tasks.

\medskip\noindent{\textit{Acknowledgments.}}
This work was partially supported by the Italian Ministries MIMIT, under project EI-TWIN n. F/310168/05/X56 CUP B29J24000680005, project ASVIN n. F/360050/01-02/X75 CUP B29J2400020000, and MUR, under projects: PNRR FAIR - Spoke 9 - WP 9.1 CUP H23C22000860006, Tech4You CUP H23C22000370006, and PRIN PINPOINT CUP H23C22000280006.

\paragraph{Competing interests.} The author(s) declare none.

\bibliographystyle{acmtrans}
\bibliography{name}

\label{lastpage}
\end{document}